\pdfoutput=1

\documentclass[11pt]{article}

\usepackage[preprint]{acl}

\usepackage{times}
\usepackage{latexsym}
\usepackage{multirow}
\usepackage{amsmath}
\usepackage{amssymb}
\usepackage{caption}
\usepackage{cuted}
\usepackage{booktabs} 
\usepackage{mathptmx}
\usepackage{tabularx}
\usepackage{array}

\usepackage[T1]{fontenc}

\usepackage[utf8]{inputenc}

\usepackage{microtype}

\usepackage{inconsolata}

\usepackage{graphicx}

\title{Writing as a Self-Organized Critical Process}

\author{Nikolay Mikhaylovskiy \\
NTR Labs / Moscow, Russia \\
Higher IT School of Tomsk State University / Tomsk, Russia \\
\texttt{nickm@ntr.ai} \\}

\begin{document}

\maketitle

\begin{abstract}
We explain autocorrelation decay power laws omnipresent in texts by self-organized criticality. Specifically, we analyze the recently released KLiCKe keystroke dataset and show that not only the final texts' autocorrelations form a manifold that adheres to a power law with a finite-size scaling, but also the text revisions generate revision-size-dependent restoring dynamics toward that manifold. Thus, human writing appears to dynamically regulate semantic correlations in a text toward a critical state.
\end{abstract}

\section{Introduction}

Whenever one turns to the statistical properties of human-written texts, power laws are omnipresent. The best known is Zipf’s law, which is a dependency between the frequency rank $r$ of a word and its frequency in a natural corpus $f(r)$ \citep{estoup1916, zipf1999psycho, Piantadosi2014ZipfsWF}:
\begin{equation}
f(r) \propto  r^{-\alpha},
\end{equation}
where $\alpha \approx 1.0$. There are several other related laws, including Heaps' law \citep{herdan1964quantitative, Heaps1978InformationRC, baayen2001word, 10.1371/journal.pone.0005372, Font-Clos_2013},  Hilberg’s law \citep{hilberg1990bekannte, e18100364} etc. Zipf's law is sometimes considered "linguistically shallow" \citep{MAN83, ManinMandelbrot}, and we focus on a different kind of power laws, autocorrelation decay laws. 

Long-range correlations in the written language have been reported using a variety of representations and observables. Early studies found slowly decaying correlations in symbolic representations of character sequences \citep{doi:10.1142/S0218348X93000083,EBELING1995233}, while later work identified long-range dependence in word-frequency and recurrence statistics \citep{doi:10.1142/S0218348X02001257,doi:10.1073/pnas.1117723109}. Most directly related to the present work, \citet{doi:10.1073/pnas.0510673103} and \citet{Mikhaylovskiy} represented texts in a distributional semantic space and found power-law semantic correlations across literary texts.

While the existence of long-range dependence in completed texts is well established, the mechanism responsible for it remains unclear. The hierarchical semantic/discourse organization proposed by  \citet{doi:10.1073/pnas.0510673103, e19070299,  Sainburg2019ParallelsIT, 10.1098/rspb.2021.2657} does not explain fine-grained correlations, and the conjecture of \citet{e19070299} has recently been specifically refuted at a fine-grained timescale by \citet{10.1162/TACL.a.785}. \citet{nakaishi2024criticalphasetransitionlarge} and \citet{mikhaylovskiystates} implicitly place human texts in a temperature-dependent continuum of texts generated by LLMs and explain critical phenomena in texts by a phase transition. As a phase transition occurs at one specific temperature, the amazing ubiquity of power law phenomena in texts remains puzzling from this viewpoint. 

A hypothesis closely related to ours was proposed by \citet{GromovMigrina2017}, who argued that natural language may constitute a self-organized critical system and interpreted complete texts as ``avalanches'' propagating over a word-cooccurrence graph.  They reported power-law distributions of text lengths in English and Russian literary works and social-media messages corpora and examined their scaling properties using both Kolmogorov-Smirnov-based power-law fits and data-collapse analysis. While this work provides an important conceptual precedent, its evidence is static: scale-free statistics of completed texts do not establish the dynamical mechanism by which a linguistic system approaches or remains near a critical state. We address this question directly by analyzing the evolution of individual texts during writing and testing whether editing dynamics restores the evolving text toward a finite-size scaling manifold.

In this paper, we propose Self-Organized Criticality (SOC) as a mechanism for critical autocorrelations behavior in texts. Specifically, we analyze the KLiCKe keystroke dataset \citep{Tian_Crossley_Van_Waes_2025}, and show that not only the final texts' autocorrelations form a manifold that adheres to a power law with finite-size scaling, but also revisions generate revision-size-dependent restoring dynamics toward that manifold. Our approach does not contradict the hierarchical organization explanation, but provides a fine-grained mechanism for the long-range correlations to appear. While phase-transition-based explanation considers texts to be in static equilibria, ours acknowledges the dynamic nature of language and explains the ubiquity of power law autocorrelations in texts.

In the further subsections of this section, we introduce the core concepts. In Section 2, we propose a self-organized criticality mechanism. In Section 3,  we describe the experiments and discuss the results.

\subsection{Autocorrelations}
\label{Autocorrelations}
 
Suppose that we have a sequence of $N$ vectors $V_i \in R^d, i \in [1,N]$. The autocorrelation function $C(r)$ is the average similarity between the vectors as a function of the lag $r=i-j$ between them. The simplest metric of vector similarity is the cosine similarity 
\begin{equation}	
d(V_i,V_j )=
\cos(V_i,V_j)=
\frac{(V_i\cdot V_j)}{||V_i|| ||V_j||},
\end{equation}
where $\cdot$ is a dot product between two vectors and $||x ||$ is a Euclidean norm of a vector. Thus,
\begin{equation}	
C(r)=\frac{1}{N-r} \sum_{i=1}^{N-r} \frac{V_i\cdot V_{i+r}}{||V_i|| ||V_{i+r}||}
\label{eq:Cr0}
\end{equation}
 
A distributional semantic model \citep{Harris01081954} such as FastText  \citep{bojanowski2016enriching} represents words by vectors. Thus, a text is transformed into a sequence of vectors, and one can calculate an autocorrelation function of the text.

\subsection{Power laws and finite-size scaling}
\label{sec:power_laws_fss}

A scale-free, power-law distribution has the asymptotic form
\begin{equation}
    C(r) \propto r^{-\gamma}.
\end{equation}
Unlike an exponential law,
\begin{equation}
    C(r) \propto \exp(-r/\xi),
\end{equation}
which is characterized by a finite correlation length $\xi$, a power law has no intrinsic characteristic scale. 

The connection between critical singularities, correlation functions, and a diverging correlation length was developed in the modern theory of critical phenomena by \citet{Fisher1964}. The observation of a power law alone is not sufficient to establish criticality. For example, broad mixtures of characteristic scales can also produce approximately power-law statistics. A more restrictive prediction of critical phenomena is the systematic dependence of the scaling regime on the size of the system \citep{Widom1965, Kadanoff1966}. While the correlation length of a theoretical infinite system diverges, in a finite system this divergence is necessarily truncated by the system size $L$, so that the effective correlation range becomes size-dependent \citep{FisherBarber1972}.

For the normalized autocorrelation profiles considered here, we examine the empirical finite-size scaling form
\begin{equation}
    Q(r,L)
    \equiv \frac{C(r,L)}{C(r_0,L)}
    \approx
    H\!\left[r\left(\frac{L_{\mathrm{ref}}}{L}\right)^D\right],
    \label{eq:fss_correlation}
\end{equation}
where $r_0$ is a fixed normalization lag, $L_{\mathrm{ref}}$ is a common reference length, and $H$ is a common scaling function over the fitted overlap range.

This form predicts approximate data collapse under horizontal rescaling \citep{FisherBarber1972}. We estimate $D$ by minimizing the discrepancy between the normalized curves. Thus, after normalization at the reference lag, the collapse uses one fitted rescaling exponent and no additional fitted vertical rescaling.

\subsection{Self-organized criticality}
\label{sec:soc}

Self-organized criticality was introduced by \citet{BakTangWiesenfeld1987} as a mechanism by which a dynamical system can spontaneously evolve towards a critical state and remain there by a feedback mechanism. This contrasts a phase transition, where the critical behavior is observed only when an external parameter, such as temperature, is sufficiently close to its critical value.

The canonical SOC example is the sandpile model \citep{BakTangWiesenfeld1987, BakTangWiesenfeld1988}. A slowly driven system accumulates a local quantity, for example, grains of sand, until a local stability threshold is exceeded. The resulting instability redistributes the accumulated quantity to neighboring degrees of freedom and may trigger further instabilities. A single perturbation can therefore initiate an avalanche whose size and duration are not fixed in advance. In the critical regime, avalanche statistics may become scale free, and its probability 
\begin{equation}
    P(S) \propto S^{-\tau},
\end{equation}
where $S$ denotes the avalanche size, with similar scaling laws for the avalanche duration and other observables.

A defining feature of the original SOC picture is the separation between slow driving and fast relaxation. This leads to the schematic dynamical cycle: 
$
    \text{slow drive}
    \;\longrightarrow\;
    \text{local instability}
    \;\longrightarrow\;
    \text{avalanche relaxation}
    \;\longrightarrow\;
    \text{continued drive}.
$
Through repeated cycles, the system approaches a state in which perturbations can propagate over a broad range of spatial and temporal scales.

This mechanism distinguishes SOC from the more general observation of critical-like statistics. A power-law distribution or long-range correlation may indicate scale invariance, but none of these observations by itself establishes that the system is self-organized. The stronger SOC claim requires a dynamical mechanism through which deviations from the critical state are compensated by the system's own evolution. Modern treatments therefore emphasize the distinction between criticality, which concerns the statistical properties of a state, and self-organized criticality, which additionally concerns the dynamics by which that state is reached and maintained \citep{Jensen1998,Pruessner2012}.

A useful representation of this distinction is to introduce a state variable $x$ measuring the displacement from a critical finite-size-scaling manifold. Self-organization requires a restoring tendency,
\begin{equation}
    \mathbb{E}\!\left[
        \Delta x \mid x
    \right]
    \simeq
    -\kappa x,
    \qquad
    \kappa > 0.
    \label{eq:soc_restoring}
\end{equation}
Such a relation provides a direct empirical route for testing self-tuning, rather than inferring it solely from power-law observables.

Consequently, a convincing empirical case for SOC is a conjunction of three kinds of evidence:
\begin{enumerate}
    \item scale-free or finite-size-scaling statistics;
    \item dynamical fluctuations or cascades spanning a broad range of scales;
    \item evidence that the system's own dynamics regulates its distance from the scaling state.
\end{enumerate}

\subsection{Writing processes and revision episodes}

Cognitive research describes writing as a recursive interaction between planning, text production, and reviewing. In the model of \citet{FlowerHayes1981}, these processes can interrupt each other and become embedded, guided by the writer’s goals. The developing text forms part of the task environment and influences subsequent decisions. Later models further distinguish the processes, control mechanisms, and cognitive resources involved in composing \citep{Hayes2012}. This perspective provides a basis for examining the feedback between the current text and the subsequent writing activity.

Keystroke logging makes the observable dynamics of writing accessible through the timing and location of insertions, deletions, and cursor movements \citep{LeijtenVanWaes2013}. However, these observations do not identify cognitive activities uniquely: reviewing may produce no textual change, and a pause may accompany several different processes. Consequently, the segmentation of writing into production bursts and revision episodes requires explicit operational definitions \citep{BaaijenEtAl2012}. \citet{ConijnEtAl2024} address this problem by extracting complete revision events, including insertions into existing text, deletions, and subsequent replacement text.

The present study examines whether observable writing activity collectively produces restoring dynamics in the semantic correlations of the evolving text. Cognitive models motivate the possibility of such feedback, while its statistical form and its relation to self-organized criticality remain an empirical question we aim to solve.

\section{Mechanism of self-organized criticality in writing}
\label{sec:proposed_soc_mechanism}

We propose that self-organized criticality in writing arises from the interaction between bursty text-production and revision activity and a scale-dependent statistical state of the evolving text. We will show that completed texts exhibit approximately power-law semantic correlations whose normalized profiles admit a finite-size collapse (Equation~\ref{eq:fss_correlation}). We use these length-dependent profiles as an empirical scaling manifold.

Writing does not approach this manifold monotonically. Text production and revision occur in intermittent bursts, or writing avalanches, which locally reorganize the text and can move its semantic correlation structure either toward or away from the scaling manifold. Let $x_t$ denote the length-conditioned distance of the evolving text from the manifold. The central dynamical hypothesis is that subsequent writing activity generates a restoring response, (\ref{eq:soc_restoring}).

The more specific SOC prediction is that this restoring tendency is coupled to activity,
\begin{equation}
    \frac{d\kappa(A)}{dA} > 0.
    \label{eq:activity_feedback}
\end{equation}
In this picture, writing avalanches are not merely sources of fluctuation. They also constitute the mechanism through which deviations from the scale-invariant state are corrected. Continued composition drives and perturbs the text, while bursts of production and revision repeatedly reorganize its semantic structure so that the evolving text remains confined near the finite-size scaling manifold.

The resulting state is dynamic rather than static: substantial writing activity continues even after the text has entered the vicinity of the manifold, but fluctuations around it become confined. In this sense, criticality characterizes the statistical organization of the text, whereas self-organization refers to the activity-dependent dynamics that maintain that organization during writing.

\section{Experiments and Results}

\subsection{Data and methods}
\label{sec:data_methods}

We use the KLiCKe corpus \citep{Tian_Crossley_Van_Waes_2025}, which contains keystroke logs and final texts for 4,992 English argumentative essays.  Each log records the timing and type of editing events, cursor position, and the corresponding text change. For analyses of completed texts, we use the final essay texts  directly. For dynamical analyses, we reconstruct the evolving text by replaying the keystroke sequence.  Of the 4,992 essays, 4,781 could be replayed exactly and were retained for trajectory-based analyses.

Analysis cohorts differ because their eligibility criteria differ. The centering comparison uses 4,976 complete-case final texts, whereas the length-binned collapse uses 4,975: one 148-word text falls below the 150-word lower boundary of the length bins. Temporal-burst analysis uses recorded events from all 4,992 essays and does not require exact text reconstruction. Trajectory analyses start from the 4,781 exact replays and apply analysis-specific eligibility and exclusion rules.

Texts are tokenized into lowercase word-like units and represented using 300-dimensional pretrained RCSLS-aligned multilingual fastText embeddings computed on Wikipedia \citep{joulin-etal-2018-loss}, but in English only. We measure correlation using centered dot products normalized by the embedding variance within the text. For a text state $t$, let $L_t$ be its word count, $\mathcal O_t$ its covered word positions, and $n_t=|\mathcal O_t|$. Define
\begin{equation}
\begin{split}
 \bar V_t&=\frac{1}{n_t}\sum_{i\in\mathcal O_t}V_{ti},\\
 Z_{ti}&=V_{ti}-\bar V_t,\qquad
 v_t=\frac{1}{n_t}\sum_{i\in\mathcal O_t}\|Z_{ti}\|^2.
\end{split}
\end{equation}
At positive lag $r$, let $\mathcal P_t(r)=\{i:i,i+r\in\mathcal O_t\}$ and $P_t(r)=|\mathcal P_t(r)|$. The excess correlation is
\begin{equation}
 C_t(r)=\frac{\sum_{i\in\mathcal P_t(r)}
                  Z_{ti}^{\mathsf T}Z_{t,i+r}}
                 {P_t(r)v_t}
          +\frac{1}{n_t-1}.
 \label{eq:Cr}
\end{equation}
The additive term subtracts the expected correlation under random permutation of the covered vectors among their positions, correcting the negative baseline induced by within-text centering. Out-of-vocabulary words retain their positions, but do not enter the mean, variance, or pair sums. Equation~\ref{eq:Cr} is used for individual texts; the corresponding pooled estimator is defined in Appendix~\ref{app:correlation-state-definition}. For completed-text length bins, we pool centered dot products and valid-pair counts across essays before normalizing by the pooled embedding variance. Individual trajectory states are scored separately. Both constructions subtract their exact within-text shuffle baseline (Appendix~\ref{app:correlation-state-definition}). We use a methodology largely following \citet{doi:10.1137/070710111} to assess power-law fitness.

Writing activity is measured from the replayed logs as the number of characters inserted or removed per unit elapsed time. Snapshot analyses use sampled text trajectories, whereas complete-revision analyses score the immediate event endpoints. Whole-essay resampling quantifies uncertainty in the reported regression estimates; the model-based tail goodness-of-fit diagnostics use a separate procedure described in Appendix~\ref{app:typing-activity-scaling}.

\begin{figure*}[t!]
    \centering
    \includegraphics[width=\textwidth]{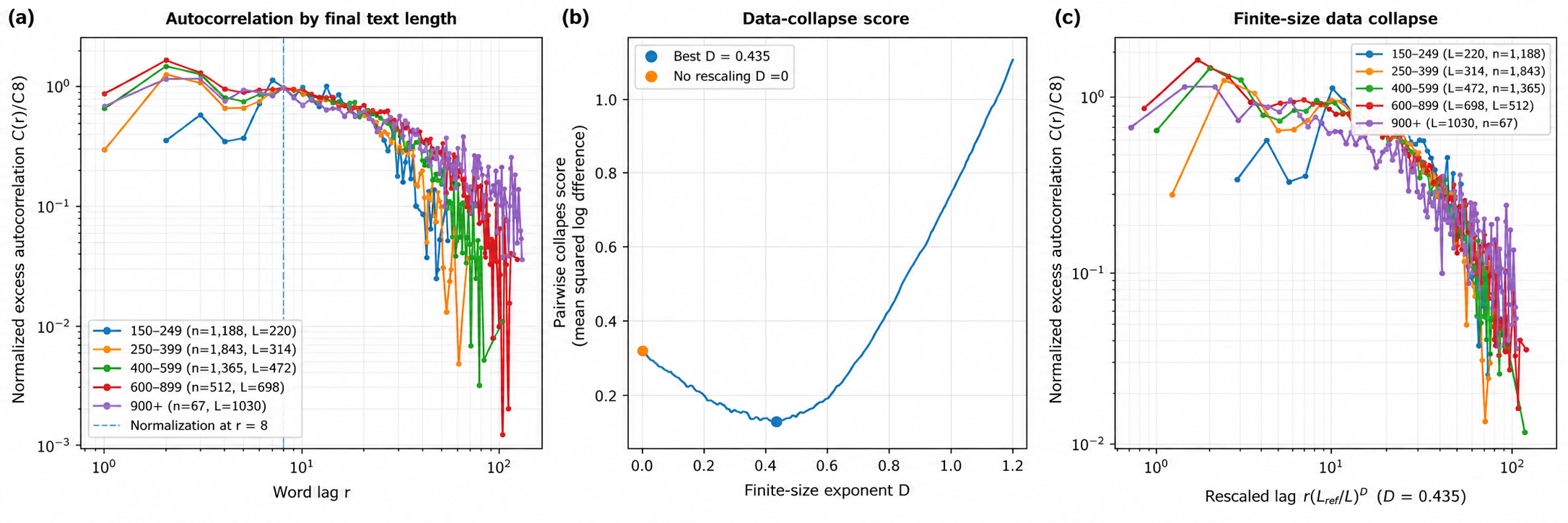}
    \caption{
    Finite-size collapse of normalized semantic autocorrelations in completed KLiCKe essays.
    \textbf{(a)} Normalized excess autocorrelation $C(r, L)/C(8, L)$ for bins of final text length.
    \textbf{(b)} Data-collapse score as a function of the horizontal rescaling exponent $D$, with the optimum at $D=0.435$.
    \textbf{(c)} The same autocorrelation curves after rescaling the lag as $r(L_{\mathrm{ref}}/L)^D$, showing improved collapse across text lengths.
    }
    \label{fig:fss}
\end{figure*}

\subsection{Completed texts exhibit power law with finite-size scaling}
\label{sec:final_fss}

We first ask whether completed essays exhibit the semantic correlations expected of a finite critical system.  Figure~\ref{fig:fss}A shows normalized excess semantic autocorrelation $Q(r,L)=C(r,L)/C(8,L)$ for final texts grouped by length. Over intermediate lags the correlations decay approximately as a power law, while at larger lags they bend downward. Crucially, this cutoff shifts systematically to larger distances for longer texts: a characteristic correlation range increases from approximately $37$ words in the shortest length bin to approximately $71$ words in the longest.

We examine the normalized scaling form in Equation~\ref{eq:fss_correlation}. Rescaling the lag as $r(L_{\mathrm{ref}}/L)^D$ substantially improves the overlap of the length-conditioned profiles (Figure~\ref{fig:fss}C). The optimal collapse is obtained at $D \simeq 0.44$; the collapse discrepancy is approximately $55\%$ smaller than for $D=0$, where no size-dependent horizontal rescaling is allowed (Figure~\ref{fig:fss}B).  Leave-one-length-bin-out estimates remain in the range $D\simeq0.36$--$0.57$, and permutation of length-bin labels gives $p\simeq0.008$.

These results provide stronger evidence than power-law fitting alone: the power-law-like regime is limited in extent, but its cutoff changes systematically with text length and the   alignment of normalized curves improves under a single horizontal-rescaling exponent. We use this family of length-dependent correlation profiles as the empirical scaling state toward which writing trajectories are compared below.

\subsection{Typing activity exhibits power-law behavior}
\label{sec:typing-activity-scaling}

We examined whether broad activity distributions persist across different definitions of an editing unit. Temporal bursts, deletion runs, and revision/production episodes capture different aspects of writing; their boundaries and size measures are not interchangeable. Several distributions admit approximately power-law tails, but the evidence varies across measures (Table~\ref{tab:writing-activity-tails}).

A \emph{temporal burst} is a maximal sequence of text-changing events separated by gaps $g_i\leq1.25m_e$, where $m_e$ is the median inter-event interval in essay $e$. The multiplier $c=1.25$ was adopted following an exploratory sweep over $c\in\{0.5,0.75,1,1.25,1.5,2\}$, comparing burst-size distributions and between-essay variability with fixed temporal thresholds; details are reported in the supplementary analysis in Appendix~\ref{app:1.25}. Its size $S$ counts events and its duration is $T=(t_{\mathrm{last}}-t_{\mathrm{first}})/m_e$. Excluding the first and last burst per essay yielded 5,270,050 avalanches across 4,992 essays. A \emph{deletion run} groups consecutive removal or cut operations; its size $W$ is the net word loss, analyzed conditional on $W>0$.

To include deletion and subsequent replacement, we also extracted \emph{revision episodes} following \citet{ConijnEtAl2024}. Revisions start with deletion or insertion into existing text; the authors' released keystroke-only support-vector model selects endpoints within structurally bounded candidates. \emph{Production episodes} are maximal nonempty sequences outside the extracted revisions. We retained the conservative complete subset, excluding episodes that could reach the recording boundary. Episode size $V$ counts gross inserted plus deleted characters, including whitespace, both components of replacements, and bulk edits.

We fitted tails of the form $p(z\mid z\geq z_{\min})\propto z^{-\alpha}$ and compared them with alternative distributions. The temporal-size fit uses 1,086 bursts with $S\geq63$, only $0.0206\%$ of the 5,270,050 primary bursts; its exponent is $3.159$. The deletion-size fit uses 9,040 runs with $W\geq6$, or $4.24\%$ of positive-word-loss runs, and gives an exponent of $2.640$. Both fits favored power laws over exponential alternatives, while a lognormal remained plausible for temporal-burst sizes. Appendix~\ref{app:large-revision-cases} examines essays contributing extreme revision episodes.

Revision-size exponents stayed near three as the lower threshold varied from 100 to 500 characters. At 100 characters, held-out essay comparisons favored the power law over an exponential, but did not clearly distinguish it from lognormal or cutoff alternatives. Production showed a possible crossover around 300--400 characters, supported by only 236 episodes from 187 essays above 300, and 116 episodes from 89 essays above 400.

Duration required a different description: a descriptive two-regime log--log CCDF fit identified a crossover at $T_\ast=36.83$ (95\% whole-essay bootstrap interval $[31.66,48.74]$). These results support approximately power-law tails in selected activity measures, without establishing a power law over their full distributions or a universal exponent. Full distribution plots and fitting diagnostics
are provided in Appendix~\ref{app:typing-activity-scaling}.

\begin{table}[t]
\centering
\small
\setlength{\tabcolsep}{3pt}
\begin{tabular}{@{}lll@{}}
\toprule
Measure & Range / threshold & Exponent $\alpha$\\
\midrule
Burst size $S$ & $S\geq63$ & $3.159$\\
Word loss $W$ & $W\geq6$ & $2.640$\\
Revision size $V$ & $V_{\min}=100$--$500$ & $\approx3$\\
Burst duration $T$ & $5\leq T\leq185.85$ & $4.742\,/\,2.706$\\
\bottomrule
\end{tabular}
\caption{\textbf{Tail behavior across writing-activity measures.} Size measures use different units: events ($S$), net words ($W$), and
gross characters ($V$). The revision entry summarizes fits across lower thresholds of 100--500 characters. Duration entries are density-equivalent exponents derived from CCDF slopes below/above $T_\ast$, rather than a single fitted power-law density.}
\label{tab:writing-activity-tails}
\end{table}

\subsection{The scaling state emerges during writing}
\label{sec:scaling_emerges}

\begin{figure}[t!]
    \centering
    \includegraphics[width=\linewidth]{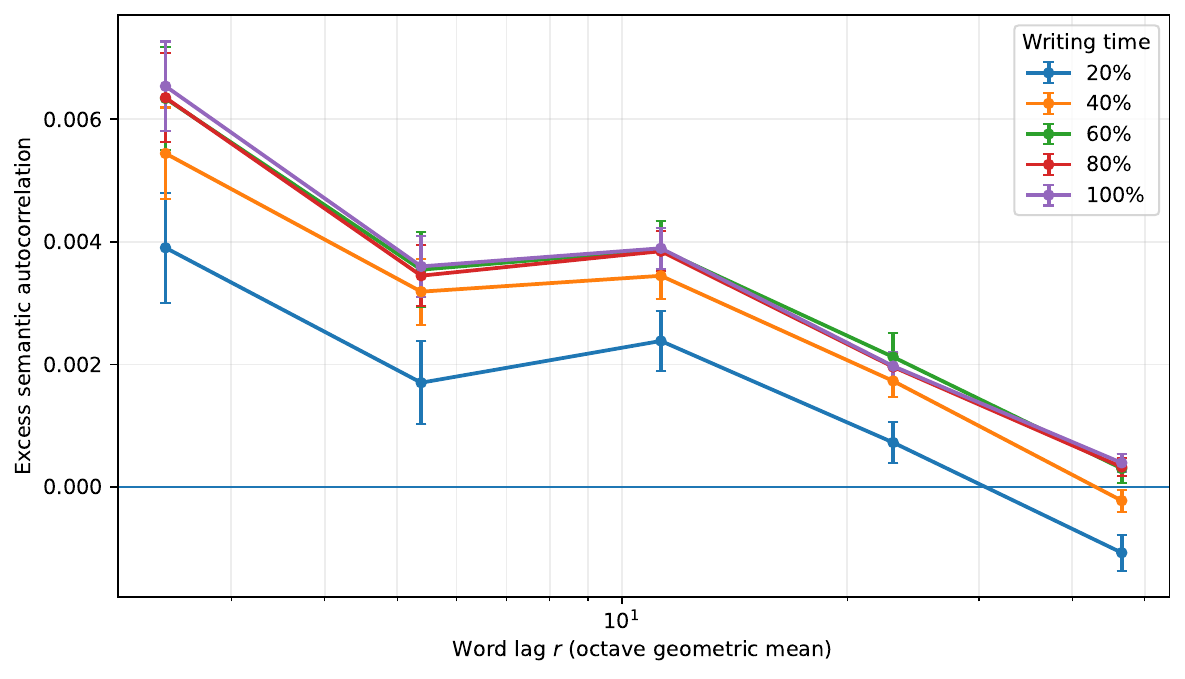}
    \caption{Emergence of long-range semantic correlations during writing.
    Curves show mean excess FastText autocorrelation at five stages of normalized writing time for a fixed cohort of 612 essays with final lengths of 250--349 words. The same essays contribute at every stage and plotted octave; error bars denote 95\% confidence intervals.}
    \label{fig:scaling-emerges}
\end{figure}

The long-range correlation structure of completed texts is not present from the beginning of composition. Figure~\ref{fig:scaling-emerges} follows a fixed cohort of essays at 20\%, 40\%, 60\%, 80\%, and 100\% of normalized writing time. Semantic correlations get progressively stronger with writing time at short and intermediate lags, while their positive range extends to larger lags. In particular, the correlation at the 32--63 word octave changes from negative at 20\% to positive by 60\% and remains positive thereafter. Thus, the mature scale-dependent correlation profile develops during composition, rather than being an immediate consequence of text length alone.

\subsection{Editing activity generates restoring dynamics}
\label{sec:restoring-dynamics}

The finite-size collapse of completed essays  (\ref{eq:fss_correlation}) does not mean by itself that the power law is dynamically produced during writing through a self-organization mechanism. We therefore replayed each keystroke log and measured how successive intermediate versions moved relative to the finite-size power law manifold inferred from completed essays. We tested this relation at two resolutions: intervals of ongoing writing and individual complete revisions.

For each text state, we sample its excess correlation at common length-rescaled lags, forming a feature vector $y\in\mathbb R^m$. A reference direction $f$ and residual scatter matrix $\Sigma$ are learned from completed essays outside the scored essay's fold. Let $B$ have orthonormal columns spanning the complement of $f$.
We define
\begin{equation}
 d_{\mathrm{crit}}
 =\left[\frac{(B^{\mathsf T}y)^{\mathsf T}
                    \Sigma^{-1}(B^{\mathsf T}y)}{m-1}\right]^{1/2}.
 \label{eq:dcrit-definition}
\end{equation}
This is the standardized residual of the fitted common correlation shape after projecting out its amplitude component. The reference is an empirical rank-one model. Its construction is separate from the normalized population collapse in Figure~\ref{fig:fss}; Appendix~\ref{app:correlation-state-definition} gives the fitting and scoring procedure.

Because the distance estimator depends on text length, we defined the standardized state
\begin{equation}
x_{ei}=\frac{d_{\mathrm{crit},ei}-m_d(L_{ei})}{s_d(L_{ei})},
\label{eq:length-adjusted-manifold-distance}
\end{equation}
where $m_d(L)$ and $s_d(L)$ are the median and robust scale of out-of-fold final-text distances at comparable lengths. The scale is the median absolute deviation across completed essays within each length bin, divided by $\Phi^{-1}(0.75)$; both quantities are interpolated across lengths. Thus, $x=0$ denotes the length-conditioned reference distance, not zero geometric distance from the manifold. For a pre/post event pair, we measured $\Delta x=x^+-x^-$. A negative starting-state coefficient indicates that higher starting states predict more negative subsequent changes, conditional on the other regressors.

For ongoing writing, the pre/post states were three-snapshot means separated by one omitted snapshot. We fitted
\begin{equation}
\begin{split}
\Delta x_{ei}={}&a_e+\beta_x x^-_{ei}+\beta_A A_{z,ei}
+\beta_L\ell_{ei}\\
&+\beta_{xA}x^-_{ei}A_{z,ei}+\varepsilon_{ei},
\end{split}
\label{eq:primary-restoring-model}
\end{equation}
where $a_e$ is an essay fixed effect, $A_z$ is inserted-plus-deleted characters per minute standardized robustly within essay, and $\ell$ is centered log word count. After excluding two essays (see Appendix~\ref{app:excluded-essays} for reasons), 106,353 windows from 4,342 essays remained. The uncorrected state coefficient was negative ($\hat\beta_x=-0.8073$), as was its interaction with activity ($\hat\beta_{xA}=-0.01247$, 95\% whole-essay bootstrap interval $[-0.01952,-0.00767]$). None of 500 within-essay circular shifts of activity produced an interaction as negative as observed ($p=0.0020$), supporting its temporal alignment with the state change.

This test does not resolve the shared estimation error. The starting estimate also enters $\Delta x$ with a minus sign: overestimating it mechanically makes the measured change more negative, potentially mimicking restoration, the effect known as regression to the mean \citep{BarnettEtAl2005}. We therefore varied the contributions of persistent word blocks to the estimator, using the same weights across successive texts to preserve uncertainty shared by retained content. The resulting covariance multiplied by parameter $\lambda$ was subtracted from the regression moments before refitting (see Appendix~\ref{app:restoration-calibration} for details and extensive discussion). This is a conditional uncertainty model, not a correction for every possible source of error.

At $\lambda=1$, the snapshot model containing the activity interaction $\beta_{xA}x^-_{ei}A_{z,ei}$ failed the requirement that corrected moments form a valid covariance structure. A reduced model omitting only that interaction, while retaining all the other effects, remained admissible. Its corrected state coefficient was $-0.8090$ with 16-word blocks (95\% interval $[-0.9398,-0.6839]$), and remained negative with 8- and 32-word blocks. All 2,000 whole-essay bootstrap draws were admissible at these settings. The negative baseline association therefore survives the specified full-covariance correction in the reduced model.

To connect feedback to individual editing avalanches, we next used the complete revisions defined above following \citet{ConijnEtAl2024}. Each observation compares the text immediately before the first native modification with that immediately after the last; bulk operations remain indivisible. Endpoint scores and their joint uncertainty were computed on this schedule. The cohort contained 292,899 scoreable revisions from 4,294 essays with at least eight eligible revisions. With gross changed-character size $V$ and centered $s=\log(1+V)-\overline{\log(1+V)}$, we fitted
\begin{equation}
\begin{split}
\Delta x_{ej}={}&c_e+\gamma_x x^-_{ej}+\gamma_Vs_{ej}
+\gamma_L\ell^-_{ej}\\
&+\gamma_{xV}x^-_{ej}s_{ej}+u_{ej}.
\end{split}
\label{eq:revision-restoring-model}
\end{equation}
All eligible sizes were included, without restriction to the fitted power-law tail. At full calibrated correction with 16-word blocks, the state--size interaction was $\hat\gamma_{xV}=-0.09140$ (95\% interval $[-0.10774,-0.07828]$). It remained negative with 8- and 32-word blocks and after adding log duration and its state interaction, with all 2,000 bootstrap draws admissible at the reported settings. Larger revisions therefore had a more negative conditional state-feedback slope under the specified error model.

The fitted slope is $g(V)=\gamma_x+\gamma_{xV}s$, so the sensitivity of the expected post-state to the pre-state is $1+g(V)$. At five and twelve changed characters, the corrected slopes were $-0.0573$ and $-0.1280$, respectively, consistent with local contraction of the conditional mean. Although 15.6\% of revisions left the estimator's normalized token sequence unchanged, a common-cohort analysis restricted to token-changing revisions showed similar size-bin feedback patterns. Those exploratory patterns were not monotonic across all bins, and regression intervals were withheld when bootstrap admissibility failed (Appendix~\ref{app:revision-size-restoration}).

\begin{figure}[t]
    \centering
    \includegraphics[
        width=\linewidth
    ]{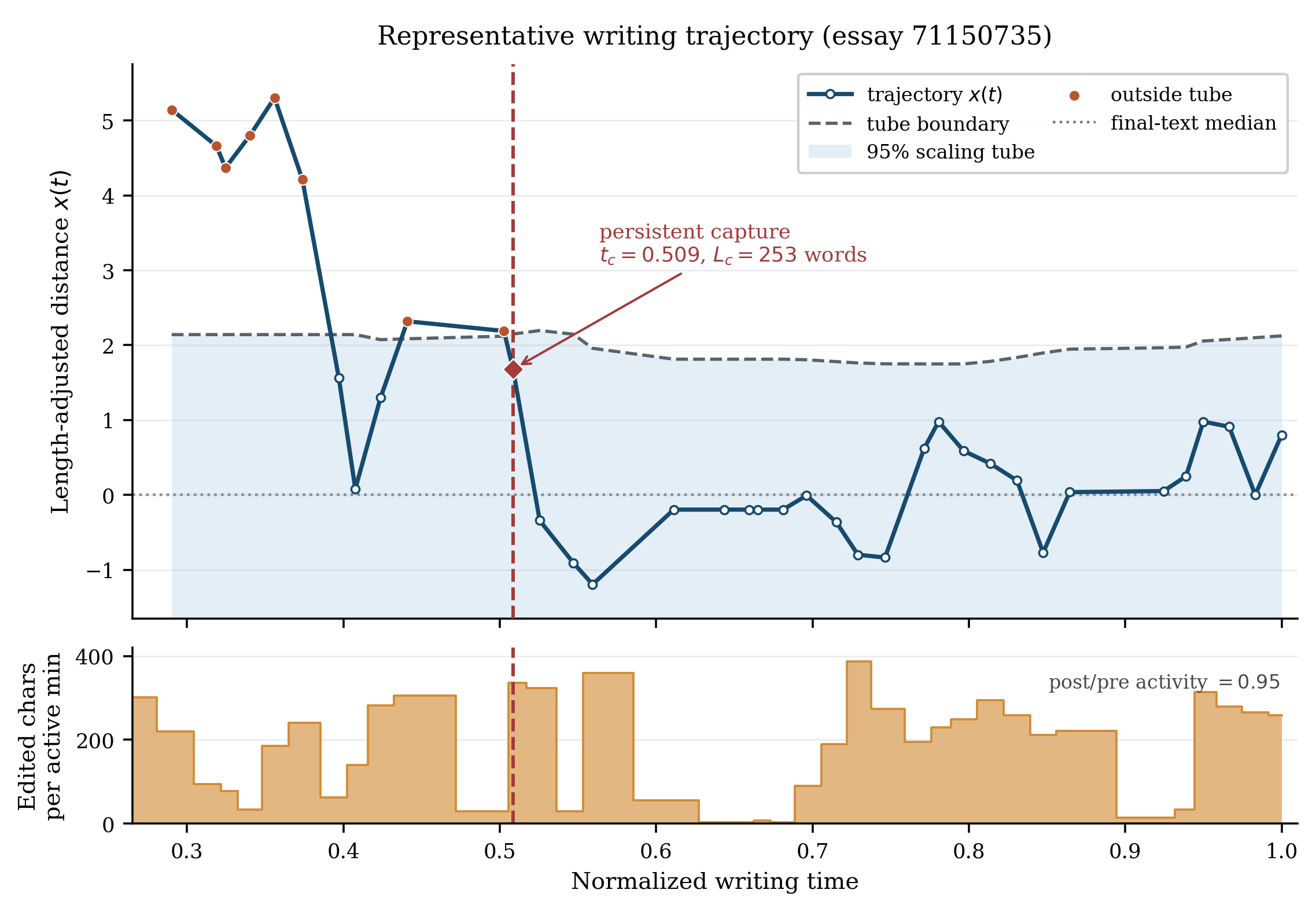}
    \caption{
Illustrative writing trajectory exhibiting approach, persistent capture, and confinement near final-text reference tube.   }
    \label{fig:capture-confinement}
\end{figure}

\subsection{Persistent occupancy of the final-text reference tube}
\label{sec:persistent-occupancy}

We additionally summarized persistent occupancy of the final-text reference tube. This analysis describes the recorded trajectories retrospectively. For eligible snapshot $i$ of essay $e$, define 
\begin{equation}
 I_{ei}=\mathbb I\!\left[
 d_{\mathrm{crit},ei}\leq q_{0.95}(L_{ei})\right],
 \label{eq:scaling-tube-indicator}
\end{equation}
where $q_{0.95}(L)$ is the interpolated length-conditioned 95th percentile of out-of-fold final-text distances. Evaluation begins at 155 words, the lower boundary of the length calibration. Indexing the eligible snapshots by $i=1,\ldots,n_e$, robust capture
is the earliest snapshot $k_e$ satisfying
\begin{equation}
\begin{split}
 I_{e k_e}=1,\qquad n_e-k_e+1\geq8,\\
 \frac{1}{n_e-k_e+1}\sum_{i=k_e}^{n_e}I_{ei}\geq0.90.
\end{split}
\label{eq:robust-capture}
\end{equation}
Thus, at least eight eligible states must remain, including the selected state, and at least 90\% must lie inside the tube. Strict capture replaces $0.90$ by $1$. These persistence criteria are distinct from the 95th percentile defining the tube boundary.

The primary observed-capture definition additionally requires at least five preceding eligible states and at least one preceding state outside the tube. We select the earliest snapshot satisfying all conditions.

These descriptive capture summaries use all 4,781 exactly replayed essays, including the two essays excluded from the restoring-dynamics regressions (Appendix E.1). Without these additional preceding-state requirements, 4,152 of 4,781 usable essays (86.8\%) met the robust capture criterion. Observed robust and strict capture occurred in 1,347 (28.2\%) and 1,307 (27.3\%) essays, respectively. Writing activity continued within the selected persistent segments
(Appendix~\ref{app:capture-definition}). Figure~\ref{fig:capture-confinement} illustrates one trajectory. Because $k_e$ is selected using subsequent occupancy, these counts are retrospective descriptions, not prospective capture probabilities.

\subsection{Ablations}
\label{subsec:robustness}

We tested whether the main results depended on corpus composition, the autocorrelation estimator, individual length bins, overlapping state measurements, or the operational definition of capture  -- overall, they don't. Detailed results are presented in Appendix~\ref{app:ablations}. 

\section{Conclusion}

Our results provide evidence for three related phenomena: power law with finite-size scaling behavior of autocorrelations in completed texts, power-law distribution of writing activities, and state-dependent dynamics that return developing texts toward the population-typical scaling relation.

The final autocorrelation profile is not present at the beginning of writing: longer-range positive correlations emerge progressively during composition. The temporal analysis provides evidence for self-organization. Deviations from the length-conditioned scaling power-law manifold were followed by compensatory changes. Activities do not produce a uniform drift in the semantic state; instead, bigger revisions strengthened the state-dependent correction. Moreover, essays tend to exhibit an observed robust capture into the scaling tube. Confinement therefore cannot be explained solely by writers stopping once the text enters the tube.

Together, these findings suggest that text editing continually perturbs the semantic organization of a developing text, while subsequent activity is statistically biased toward reducing departures from the population typical state. Writers need not represent this state explicitly or optimize an autocorrelation function. Decisions about what to add, remove, clarify, or connect may collectively generate a macroscopic restoring tendency. The resulting organization is maintained amid continued fluctuation rather than reached only when activity ceases. Accordingly, we regard self-organized criticality as a hypothesis supported by multiple converging statistical signatures.

\section*{Acknowledgements}

\section*{Limitations}

This study provides evidence for dynamical regulation toward an empirical scaling state, but it does not uniquely identify self-organized criticality in the strict statistical-physics sense.

The finite-size analysis is based on five length bins, of which the $900+$-word bin contains only $67$--$69$ essays, depending on the analysis-specific inclusion criteria. The estimated collapse parameter varies from $0.355$ to $0.570$ when individual bins are omitted. The exact length-label permutation test is correspondingly limited to $5!=120$ assignments. We therefore treat $D$ as an empirical rescaling parameter, not as a precisely estimated critical exponent. The observed scaling range is also short, particularly in the shorter essays, and asymptotic behavior cannot be established from these text lengths.

The semantic observable depends on pretrained FastText embeddings. These embeddings are static, represent a word type independently of its local sense, and omit words not covered by the embedding vocabulary. Although embedding coverage is high and out-of-vocabulary positions are retained as gaps, the resulting autocorrelation measures lexical-semantic similarity rather than the full syntactic, discourse, or pragmatic state of a text. The extreme long-range tail is also sensitive to centering and aggregation. Replication with contextual embeddings, alternative semantic representations, tokenizations, and distance measures is necessary before the inferred manifold can be considered representation independent.

The scaling manifold is learned from completed texts and is therefore an empirical reference state rather than a theoretically specified critical surface. Cross-fitting ensures that an essay is evaluated against a manifold and length calibration estimated without that essay, reducing direct information leakage. It does not establish that the final-text distribution is optimal, universal, or critical. Because the same representational family is used to construct the manifold and measure trajectory distance, unmodeled features of that representation may influence both the target and the observed restoring dynamics.

The trajectory analysis is observational. Writing activity, text state, task difficulty, and author behavior are jointly determined, and neither essay fixed effects nor the circular-shift null constitutes experimental randomization of activity. A negative coefficient on the current displacement can also be strengthened by regression to the mean, measurement error, and bounded state space. Non-overlapping state windows and the activity-alignment null reduce these concerns, particularly for the interaction $\beta_{xA}$, but they do not eliminate them. A causal claim that activity \emph{produces} restoring dynamics would require an intervention that changes revision activity independently of the text's current state.

Capture is a retrospective occupancy criterion whose onset is selected using the subsequent trajectory. Its incidence depends on the reference tube, eligible text lengths, snapshot schedule, the persistence threshold, and the required observations before and after onset. 28.2\% of the essays satisfied the robust definition, compared with 86.8\% when the preceding-state requirements were omitted. Assessing persistence beyond a null expectation would require a prospective criterion or surrogates evaluated with the same complete selection procedure.

Progress through a writing session is correlated with text length. The fixed-cohort analysis rules out changing cohort composition, while the length-standardized trajectory coordinate removes the median and robust scale expected at a given length. Nevertheless, these adjustments may not remove all mechanical effects of having more word pairs, broader discourse structure, or more stable correlation estimates in longer snapshots. The results concern population-average behavior and do not imply that every individual edit, or even every essay, moves monotonically toward the manifold.

Revision boundaries are inferred using an adapted keystroke-only endpoint model \citep{ConijnEtAl2024}, whose transfer to KLiCKe has not been independently validated against annotated boundaries. Endpoint errors can merge revisions with subsequent production or truncate longer editing sequences, changing both gross episode size and the text used to measure the post-state. Boundary uncertainty can therefore affect the estimated size distributions and the association between revision size and state change. Our whole-essay bootstrap and shared-error calibration condition on the extracted boundaries and do not propagate this uncertainty. The results concern these operationally defined episodes; their correspondence to cognitive revision units remains uncertain. Targeted annotation across episode sizes and sensitivity analyses using plausible alternative endpoints would help assess robustness to segmentation.

Finally, the data come from one English-language essay-writing setting and a restricted range of text lengths and participant behaviors. Keystroke logs record observable editing operations but not planning, attention, intention, or off-screen composition. Replay and event-classification errors may also affect the reconstructed trajectories and avalanche boundaries. The findings may not transfer to other languages, genres, professional or collaborative writing, speech, very long documents, or writing performed with substantial LLM assistance. Replication across tasks and controlled manipulation of revision activity are required to determine which aspects of the observed restoring dynamics generalize beyond this corpus.

\bibliography{anthology,custom}

\appendix

\section{Sensitivity analysis using calibrated estimation-error covariance}

\subsection{Correlation estimator and empirical state distance}
\label{app:correlation-state-definition}

\paragraph{Pooling completed texts.}
Using the quantities in Equation~\ref{eq:Cr}, write
$S_e(r)=\sum_{i\in\mathcal P_e(r)}Z_{ei}^{\mathsf T}Z_{e,i+r}$.
For a length bin containing essays $\mathcal E$, the pooled variance and excess correlation are
\begin{align}
 v_{\mathcal E}&=\frac{\sum_{e\in\mathcal E}n_ev_e}
                         {\sum_{e\in\mathcal E}n_e},\\
 C_{\mathcal E}(r)&=
 \frac{\sum_{e\in\mathcal E}
          [S_e(r)+P_e(r)v_e/(n_e-1)]}
      {v_{\mathcal E}\sum_{e\in\mathcal E}P_e(r)}.
 \label{eq:pooled-excess-correlation}
\end{align}
Thus, the pooling weights depend on valid-pair counts and embeddingvariances; this estimator is not an equal-essay average of $C_e(r)$. The baseline follows from the expected centered dot product $-v_e/(n_e-1)$ for distinct covered positions under within-essay permutation. The plotted population profiles are subsequently normalized as $C_{\mathcal E}(r)/C_{\mathcal E}(8)$. The saved completed-text run used lags $1$--$128$, at least 20 valid pairs at every lag, and embedding coverage of at least $0.60$.

\paragraph{Features and cross-fitted reference.}
For a training fold complement, let $L_{\mathrm{ref}}$ be the median training-text length and let $q_1,\ldots,q_m$ be the configured reference lags. For each candidate exponent $D$, a text of length $L_t$ is represented by
\begin{equation}
 y_{tj}(D)=\mathcal I_t\!\left(
      q_j(L_t/L_{\mathrm{ref}})^D\right),\qquad j=1,\ldots,m,
 \label{eq:manifold-features}
\end{equation}
where $\mathcal I_t$ interpolates $C_t(r)$ linearly in $\log r$. Correlation values themselves are neither logged nor divided by $C_t(8)$. Negative excess correlations are retained. Only finite lag estimates meeting the configured minimum pair count enter interpolation; at least three usable lags are required, extrapolation is forbidden, and all $m$ target coordinates must be available. Nonfinite or numerically zero feature vectors are rejected.

Stacking eligible training features as rows gives $Y(D)$. The exponent is selected on the configured grid by minimizing the fraction of squared feature magnitude left unexplained by a rank-one approximation,
\begin{equation}
 J(D)=1-\frac{\sigma_1(Y(D))^2}{\|Y(D)\|_F^2},
 \label{eq:manifold-fit-objective}
\end{equation}
subject to the configured minimum fraction of representable training texts and at least ten valid training rows. Here $\sigma_1$ is the largest singular value. The unit reference vector $f$ is the leading right singular vector at the selected $D$. The fitting is uncentered in feature space: no training feature mean is subtracted. For a scored text, the fitted amplitude is $a=f^{\mathsf T}y$ and the residual is $y-af$. Amplitudes are unrestricted, so the fitted reference is the linear span of $f$.

Let $B^{\mathsf T}B=I$ and $B^{\mathsf T}f=0$. The residual coordinates are $u=B^{\mathsf T}y$. From the $N$ training residuals we form the uncentered second-moment matrix
\begin{equation}
 S_u=\frac{1}{N}\sum_{e=1}^{N}u_eu_e^{\mathsf T}.
\end{equation}
We shrink $S_u$ toward its diagonal and add a ridge:
\begin{equation}
 \Sigma_0=(1-\rho)S_u+\rho\operatorname{diag}(S_u),\qquad
 \Sigma=\Sigma_0+\epsilon s I,
 \label{eq:reference-scatter}
\end{equation}
where $s=\operatorname{tr}(\Sigma_0)/(m-1)$ (with a fallback of $s=1$ if this scale is nonpositive or nonfinite). Whitening uses a symmetric inverse square root with eigenvalues floored at $\max(\epsilon s,10^{-12})$ before inversion. Equation~\ref{eq:dcrit-definition} is the resulting whitened residual norm divided by $\sqrt{m-1}$. This reference scatter describes variation among training-text residuals; it is distinct from the state-estimation error covariance calibrated for the restoring-dynamics regressions.

Essays are assigned to fixed folds by a deterministic hash of essay ID. The exponent, reference shape, reference length, and residual metric are fitted using completed essays outside each held-out fold and then kept fixed for every scored state of an essay in that fold. Thus, the scorer's fold-specific exponents are estimated by Equation~\ref{eq:manifold-fit-objective}; they are not fixed to the population-collapse estimate $D=0.435$ in Figure~\ref{fig:fss}. Length standardization subsequently uses the saved length-conditioned table of out-of-fold final-text distances. The revision-endpoint analysis uses the same saved reference and length table as the snapshot analysis and requires both endpoints to lie within the length-calibration support.

\subsection{Why a negative state--change association needs a sensitivity check}
\label{app:restoration-calibration}

Our restoring-dynamics regressions relate a starting state $x^-$ to its subsequent change $\Delta x=x^+-x^-$. The starting estimate occurs on both sides of this relation, with opposite signs. An overestimate of $x^-$ therefore raises the predictor and lowers the measured change, potentially creating an apparent restoring association \citep{BarnettEtAl2005}. For example, if the underlying state remains at 1, estimated endpoint pairs $(1.2,1.0)$ and $(0.8,1.0)$ produce changes of $-0.2$ and $+0.2$. 

Write the observed endpoints as $x^-=X^-+\eta^-$ and $x^+=X^++\eta^+$, where $X$ denotes an underlying state and $\eta$ estimation error. If errors have zero conditional means and are uncorrelated with the underlying states, then
\begin{equation}
\begin{split}
\operatorname{Cov}(x^-,\Delta x)
={}&\operatorname{Cov}(X^-,\Delta X)\\
&+\operatorname{Cov}(\eta^-,\eta^+)
-\operatorname{Var}(\eta^-).
\end{split}
\label{eq:shared-error-decomposition}
\end{equation}
The last two terms are the contribution to the observed association from estimation error. Independent endpoint errors yield a negative contribution. Identical endpoint errors cancel from the change. Because adjacent texts share content, their error covariance cannot generally be assumed to be zero.

\subsection{What uncertainty is being calibrated?}

The text scorer is deterministic: repeated evaluation of the same token sequence with the same reference gives the same result. Here, estimation uncertainty means uncertainty in interpreting a statistic of a finite text as an indicator of an underlying text state.

This uncertainty calibration is separate from the length calibration used to define $x$. The fitted manifold, length reference, and native revision boundaries remain fixed during uncertainty calibration.

\subsection{Synchronized weighting of persistent word blocks}

For each essay and block length $b\in\{8,16,32\}$, we generated $B=500$ joint replicates of the score trajectory. In the revision-endpoint implementation, word anchors are assigned to persistent blocks. Each replicate assigns independent positive exponential weights, with mean and variance one, to these blocks. Those weights alter the blocks' contributions to the feature calculation; they do not rearrange the text or generate new editing histories. Weighted features are transformed through the fixed reference model and standardized with the existing length calibration.

Each replicate uses one set of weights throughout an essay. Retained blocks therefore carry the same weights across endpoint states. This preserves dependence arising from shared content, unlike independently perturbing every endpoint. Identical normalized token sequences reuse the same score replicate, including when the same sequence is reached again after intervening edits. A revision leaving that sequence unchanged consequently has zero estimated change in every replicate.

For each endpoint $t$, the replicate departures are
\begin{equation}
u_{et}^{(r)}=x_{et}^{(r)}-\frac{1}{B}\sum_{q=1}^{B}x_{et}^{(q)}.
\end{equation}
Their variation estimates uncertainty in an endpoint, while their co-variation estimates uncertainty shared across endpoints. Centering at the replicate mean estimates covariance; it does not remove a systematic difference between that mean and the original score. The three block lengths assess sensitivity to the assumed extent of dependence within the text.

For the snapshot-window analysis, trajectory perturbations are propagated through the same pre/post averaging used for the observed data. For complete revisions, covariance is recalibrated at the exact first/last modification boundaries, rather than interpolated from snapshot uncertainty. The two analyses therefore use their own observation schedules.

\subsection{Subtracting the calibrated contribution and refitting}

In a regression with only a starting-state predictor and an intercept, the correction has the transparent form
\begin{equation}
\beta_{\lambda}=\frac{C_{\mathrm{obs}}-\lambda C_{\mathrm{err}}}
                         {V_{\mathrm{obs}}-\lambda V_{\mathrm{err}}},
\label{eq:scalar-error-correction}
\end{equation}
where $C_{\mathrm{obs}}=\operatorname{Cov}(x^-,\Delta x)$, $V_{\mathrm{obs}}=\operatorname{Var}(x^-)$, $C_{\mathrm{err}}=\operatorname{Cov}(\eta^-,\eta^+)-\operatorname{Var}(\eta^-)$, and $V_{\mathrm{err}}=\operatorname{Var}(\eta^-)$. Both the numerator and denominator are adjusted. A correction can therefore make a slope more negative or less negative.

The implemented models extend this operation to several predictors and essay fixed effects. Let $W_e=[Z_e,y_e]$ contain the observed regressors and response $y_e=\Delta x_e$, and let $M_e=I-\mathbf{1}\mathbf{1}^{\mathsf T}/n_e$ remove the mean over the $n_e$ observations in essay $e$. The observed joint cross-product matrix is
\begin{equation}
G_{\mathrm{obs}}=\sum_e W_e^{\mathsf T}M_eW_e.
\end{equation}
For replicate $r$, construct the corresponding error matrix $N_e^{(r)}$. Its starting-state column is $u_e^-$ and its response column is $u_e^+-u_e^-$. For a state--size interaction $x^-s$, the error column is $u_e^-s$; analogous propagation applies to state--activity and state--duration interactions. Size, activity, length, and duration themselves are treated as fixed observed modifiers. We multiply by each modifier before removing essay means.

The calibrated error cross-product matrix and corrected regression are
\begin{align}
G_{\mathrm{err}}&=\sum_e\frac{1}{B-1}\sum_{r=1}^{B}
       (N_e^{(r)})^{\mathsf T}M_eN_e^{(r)},\\
G_\lambda&=G_{\mathrm{obs}}-\lambda G_{\mathrm{err}},\\
\hat{\boldsymbol\beta}_\lambda
  &=G_{\lambda,ZZ}^{-1}G_{\lambda,Zy}.
\label{eq:shared-error-correction}
\end{align}
The subscripts identify the predictor--predictor and predictor--response blocks. This construction includes shared endpoints and dependence across observations induced by essay demeaning. Its interpretation requires error to be orthogonal to the latent joint variables, conditional on the fixed modifiers, and requires the resampling covariance to approximate that error covariance.

Two uses of ``full'' should be distinguished. \emph{Full covariance} retains the off-diagonal error terms, including those between predictors and the response. \emph{Full calibrated correction}, $\lambda=1$, subtracts the entire estimated matrix; $\lambda=0$ gives the uncorrected fit, and intermediate values subtract a fraction. Neither usage implies that all sources of uncertainty have been identified or removed.

\subsection{Admissibility and uncertainty of the fitted results}

Subtraction must leave a valid joint moment structure: the corrected predictor matrix must be positive definite and numerically well conditioned, and the corrected joint matrix must pass the positivity check. Otherwise, some linear combination of the variables would have an impossible or degenerate variance, or the coefficients would not be stably estimable. Such a fit is reported as inadmissible; negative eigenvalues are not clipped and no ridge term is added to force a result.

In the snapshot analysis, the model containing the state--activity interaction failed this requirement before $\lambda=1$ for all three block lengths. The reduced sensitivity model removed only that interaction, retaining starting state, activity, length, and essay effects. Its full-correction estimates were admissible. This supports a conclusion about the baseline association in that reduced model; it does not establish a fully corrected activity interaction. The complete-revision size-interaction models are distinct with separately calibrated endpoint errors and were admissible at the reported full-correction settings.

We used 2,000 whole-essay bootstrap samples to quantify sampling variation in the corrected fits. Each selected essay contributes its full observed and calibrated moment matrices, including repeated copies when selected more than once, and the regression is refitted. This preserves within-essay dependence. These draws differ from the 500 word-block replicates used to estimate error covariance: the inferential bootstrap holds the reference, event boundaries, centering constants, and calibrated essay-level error moments fixed. It does not propagate uncertainty in estimating those quantities. Reported regression intervals use settings for which all bootstrap draws were admissible; size-bin slope intervals were withheld when draws failed.

\subsection{Scope of the inference}

Block weighting measures sensitivity to text composition. Its interpretation as measurement-error covariance is not independently validated by repeated observations of a known latent state. Separating process variation from observation uncertainty can itself be difficult \citep{AugerMetheEtAl2016}. Our analysis consequently asks whether the association survives subtraction of a specified, empirically calibrated covariance contribution. It does not identify the true trajectory or correct nonlinear estimator bias, uncertainty in the fitted reference, or dynamic fixed-effects bias \citep{Nickell1981}. The surviving negative associations provide conditional robustness evidence, rather than proof of a causal restoring mechanism.

\section{Ablations}
\label{app:ablations}

\subsection{Alternative explanations and surrogate tests}
\label{subsec:alternative_explanations}

A power-law-like autocorrelation curve and even finite-size rescaling are not, by themselves, sufficient to identify critical dynamics. Similar signatures can arise from  hierarchical organization. We therefore repeated the autocorrelation analysis under several structure-preserving surrogates. The analysis included $4{,}975$ completed texts and $30$ surrogate realizations per condition.

For each essay, we constructed a binary segmentation tree by recursively selecting the split that maximized the cosine separation between the mean essay-centered embeddings of the two children. Segmentation stopped at leaves of approximately $64$ words. Surrogate essays were generated by independently exchanging the left and right children of each internal node while preserving the inferred groups and the word order within leaves. These hierarchy-preserving surrogates closely reproduced the observed scaling: their median collapse exponent was $D^\ast=0.380$ ($\mathrm{IQR}=0.360$--$0.400$), their median collapse improvement was $0.670$ ($\mathrm{IQR}=0.637$--$0.689$), and their median fixed-range exponent was $1.318$, compared with $1.368$ in the original texts. Their median log-RMSE from the original curves was $0.327$ ($\mathrm{IQR}=0.319$--$0.338$).

As a matched control, we retained the same binary-tree topology, leaf-size multiset, and leaf-depth multiset but randomly reassigned leaf sizes to surface positions before applying the same branch exchanges. This non-semantic hierarchy also retained much of the scaling, although less accurately: median $D^\ast=0.425$, median $I_{\mathrm{coll}}=0.403$, median exponent $1.237$, and median log-RMSE $0.388$. The semantic hierarchy produced the smaller shape error in all $30$ paired realizations. Semantic boundary placement therefore improves the reconstruction, but generic nested chunk geometry already accounts for a substantial part of the observed scaling.

These results do not rule out critical dynamics. They do show, however, that neither the apparent algebraic decay nor its finite-size collapse uniquely requires a SOC mechanism. The stronger evidence for self-organization comes instead from the independent temporal results: activity-dependent restoring dynamics and capture and confinement near the scaling manifold.

\subsection{Centering and aggregation}
Semantic autocorrelation was estimated on a common complete-case set of $4{,}976$ essays using three aggregation schemes. The primary estimator centers embeddings within each essay, pools word pairs by lag, normalizes by the pooled zero-lag variance, and subtracts the exact finite-size shuffled-centering baseline. We compared it with an equal-essay average of individually normalized autocorrelations and with a globally centered, pair-pooled estimator. The two essay-centered estimators produced the same qualitative short- and intermediate-range decay, although the exact end of the positive excess tail was estimator dependent: the equal-essay and pair-pooled curves first became non-positive at lags $56$ and $69$, respectively. The globally centered curve remained positive throughout the measured range because global centering retains persistent differences in topic and style between essays. We therefore do not interpret that long tail as within-text correlation and use essay-centered estimates throughout the primary analysis. The agreement of the two essay-centered estimators supports the qualitative decay pattern, while their different zero crossings caution against treating any individual cutoff as precisely determined.

\subsection{Finite-size collapse}
The collapse analysis used five final-length bins containing $4{,}975$ essays. After normalizing each profile by $C(8,L)$, rescaling the lag as $r/L^D$ gave $D=0.435$ and reduced the pairwise log-curve mismatch by $54.5\%$ relative to $D=0$. This result was not determined by a single length bin: leave-one-bin-out estimates ranged from $0.355$ to $0.570$, with median $0.435$. Under all $120$ exact permutations of the five observed length labels, only one assignment produced a collapse score at least as small as the observed one ($p=1/120=0.0083$). These checks support a systematic dependence of the correlation range on text length, although the leave-one-bin-out range also shows that $D$ should be regarded as a descriptive collapse parameter rather than a precisely estimated critical exponent.

\subsection{Fixed-cohort evolution}
The emergence of longer-range correlations during writing was recomputed on a fixed cohort of $612$ essays observed at every progress checkpoint. This prevents the apparent temporal pattern from being generated by different essays entering the analysis at different stages. In the $32$--$63$-word lag octave, the mean excess correlation changed from $-0.001074\pm0.000294$ at $20\%$ of writing progress to $0.000393\pm0.000151$ in the final texts, where the uncertainties are $95\%$ confidence half-widths. The estimate became positive by the $60\%$ checkpoint and remained positive thereafter. The same fixed essays therefore develop a longer positive correlation range as their texts evolve. This comparison controls cohort composition, but not the concurrent increase in text length; the latter is addressed by the length-conditioned manifold and is retained as a limitation below.

\subsection{Sensitivity of temporal segmentation}
\label{app:1.25}
Before adopting $\delta_e=1.25m_e$, we examined $\delta_e=cm_e$ for $c\in\{0.5,0.75,1,1.25,1.5,2\}$ in all 4,992 essays, together with fixed thresholds of 100, 150, and 200\,ms. The exploratory analysis compared pooled and equal-essay burst-size CCDFs, local slope diagnostics, and between-essay variability in connectivity and mean burst size. At $c=1.25$, the median fraction of connected event pairs was 0.660, similar to 0.650 at 200\,ms, while its between-essay interquartile range was much smaller (0.039 versus 0.312). The standard deviation across essays of log mean burst size was likewise smaller (0.096 versus 0.533). These comparisons motivated cadence-relative segmentation and the working choice $c=1.25$. The choice was exploratory; these diagnostics do not establish invariance of the subsequently fitted power-law exponent or duration crossover to the segmentation threshold.

\subsection{Restoring dynamics}

The primary snapshot-window model (Eq.~\ref{eq:primary-restoring-model}) used $106{,}353$ windows from $4{,}342$ essays after excluding essays \texttt{41766081} and \texttt{65825701}
(Appendix~\ref{app:excluded-essays}). Pre- and post-states were averages of three snapshots, with one omitted snapshot separating the two groups. Whole-essay bootstrap resampling gave an uncorrected state coefficient of $\hat\beta_x=-0.8073$ (95\% interval $[-0.8412,-0.7589]$) and a state--activity interaction of $\hat\beta_{xA}=-0.01247$ ($[-0.01952,-0.00767]$), consistent with the estimates reported in the main text and Appendix~\ref{app:restoration-supporting-results}.

Activity alignment was assessed on this revised cohort using $500$ within-essay circular shifts of the activity series. These shifts retain each essay's activity values and their cyclic ordering while changing their alignment with state transitions. None produced an interaction as negative as observed, giving a one-sided Monte Carlo $p$-value of $1/501=0.0020$. 

Although the pre- and post-state snapshot groups do not overlap, the starting estimate still enters both the predictor and the change $\Delta x=x^+-x^-$. The covariance sensitivity analysis in Appendix~\ref{app:restoration-calibration} addresses this coupling under the specified uncertainty model. At full calibrated covariance subtraction ($\lambda=1$), the model containing the activity interaction was inadmissible. The reduced model, omitting only that interaction, remained admissible and retained a negative state coefficient: $-0.8090$ with 16-word blocks (95\% whole-essay bootstrap interval $[-0.9398,-0.6839]$). The coefficient also remained negative with 8- and 32-word blocks, with all $2{,}000$ bootstrap draws admissible at each setting. Thus, the negative state--change association survives the specified correction in the reduced model; the activity interaction is supported by the uncorrected fit and the temporal-alignment comparison.

\subsection{Capture definition and retrospective persistence}
\label{app:capture-definition}

The observed-capture analysis in Section~\ref{sec:persistent-occupancy} requires at least five preceding eligible states, including an outside state, and at least eight states from the selected onset through the end of the eligible sequence. Requiring 90\% occupancy identified 1,347 of 4,781 essays (28.17\%); requiring uninterrupted occupancy identified 1,307 (27.34\%). Strict capture may be identified later than robust capture. Consequently, the 97.0\% overlap between qualifying essays does not estimate the probability of remaining inside the tube after the robust onset.

The earlier-outside requirement prevents an always-inside eligible trajectory from qualifying as observed capture. A trajectory that begins inside, subsequently leaves, and later meets the persistence criterion can still qualify. The broader robust definition omits both preceding-state requirements and identifies 4,152 essays (86.8\%). These incidence comparisons describe sensitivity to the operational definition.

Writing continued within the selected persistent segments: the median post/pre activity ratio was $0.828$ (interquartile range $[0.652,1.001]$). This conditional summary documents ongoing activity among captured essays. It does not remove the selection induced by choosing onset from future occupancy. We therefore report persistence and activity descriptively. 

\section{Additional coefficient and size-bin summaries}
\label{app:restoration-supporting-results}

The snapshot-window cohort excludes essays 41766081 and 65825701 for the previously documented reasons. The uncorrected state coefficient was $-0.8073$ (95\% whole-essay bootstrap interval $[-0.8412,-0.7589]$). At full calibrated covariance correction in the reduced model, the state coefficients for 8-, 16-, and 32-word blocks were $-0.7449$, $-0.8090$, and $-0.9215$, with intervals $[-0.8484,-0.6444]$, $[-0.9398,-0.6839]$, and $[-1.0887,-0.7648]$, respectively. All 2,000 essay-bootstrap draws were admissible at each setting.

For complete revisions, duration adjustment adds centered $\log(1+\mathrm{duration\ in\ seconds})$ and its interaction with the starting state. At 16-word blocks and full correction, the state--size interaction was $-0.11564$ (95\% interval $[-0.13745,-0.08205]$). The corresponding interaction remained negative at the other two block lengths, with all 2,000 bootstrap draws admissible at these settings.

The common cohort for comparing all and token-changing revisions contained 4,241 essays. Average improvement in squared reference distance was identified for the 101--300-character bin. The two largest bins contained 288 and 51 revisions, respectively, and had wide intervals. These event-weighted marginal summaries differ from the conditional regression slopes; their full decomposition and admissibility checks are reported in Appendix~\ref{app:revision-size-restoration}.

\begin{figure*}[t]
    \centering
    \includegraphics[width=\textwidth]{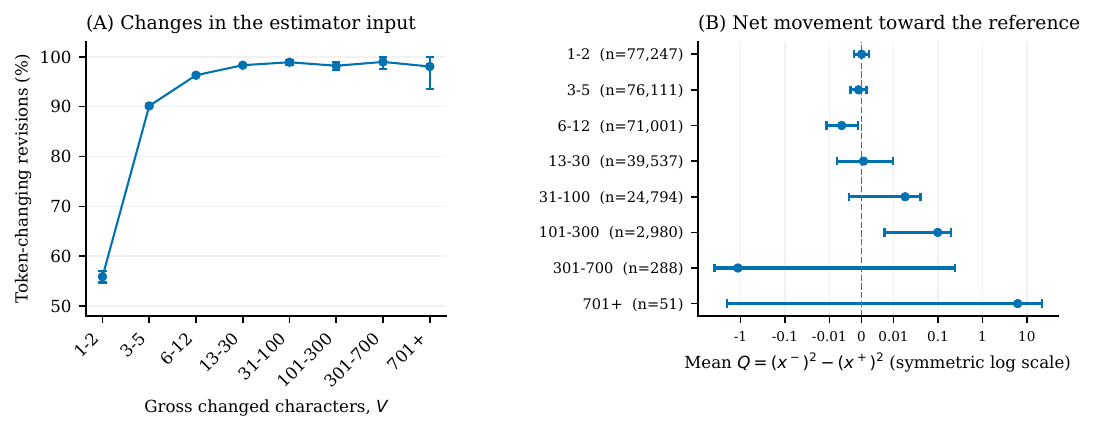}
    \caption{\textbf{Estimator changes and net restoration across revision sizes.}
    Results use the common 4,241-essay cohort and retain bulk edits. \textbf{(A)} Fraction of revisions changing the estimator's normalized token sequence. \textbf{(B)} Event-weighted mean $Q=(x^-)^2-(x^+)^2$, corrected by subtracting the calibrated pre-minus-post error-variance difference ($\lambda=1$, 16-word blocks). Positive values indicate average movement toward the length-conditioned reference. Bars are pointwise 95\% whole-essay bootstrap intervals; bin counts are shown. The horizontal axis is symmetric logarithmic, linear within $|Q|\leq0.01$, to display both the small central-bin effects and the uncertain extreme-bin estimates. These marginal means are distinct from covariate-adjusted feedback slopes.}
    \label{fig:revision-size-restoration-checks}
\end{figure*}

\section{Revision-size checks and estimator-neutral revisions}
\label{app:revision-size-restoration}

The size-bin comparison retained 4,241 essays with at least eight token-changing revisions, comprising 292,009 complete revisions, of which 246,801 changed the estimator input. The remaining 45,208 revisions (15.48\%) left the normalized token sequence unchanged and contributed zero to $Q=(x^-)^2-(x^+)^2$. This differs slightly from the 15.63\% unchanged fraction in the full endpoint cohort because of the common-cohort restriction. Token change refers to the estimator input, not an independent judgment of meaning.

For each size bin, the marginal identity $E[Q]=\Pr(\mathrm{change})E[Q\mid\mathrm{change}]$ held exactly in the saved decomposition, including defined paired bootstrap draws. It does not apply to regression slopes. At 16-word blocks and $\lambda=1$, corrected slope point estimates for all and token-changing revisions were similar, for example $-0.6313$ and $-0.6400$ in the 101--300-character bin. However, four and five of 2,000 bootstrap draws, respectively, failed admissibility in the pooled size-bin regressions. Their slope intervals were therefore withheld rather than computed after discarding failed draws. The separately estimated marginal means of $Q$ remained available.

\section{Empirical audit of influential essays and large revision episodes}
\label{app:essay-audit}

\subsection{Two essays excluded from restoring-dynamics regressions}
\label{app:excluded-essays}

Inspection of regression influence identified essays \texttt{41766081} and \texttt{65825701} as unusual contributors to the state--activity interaction. We audited their raw keystroke logs, reconstructed text trajectories, saved snapshots, and activity normalization. Strict replay required no repairs, and all 40 and 51 saved snapshots, respectively, matched the reconstructed texts and associated timing and activity records. The concerns therefore arose from the recorded behavior and its interaction with the analysis, rather than a detected replay error.

Closer inspection revealed that most of the time on these essays was spent on non-semantic typing/editing activity. 

In essay \texttt{41766081}, the reconstructed text first equaled its final version 8.545 minutes after the first text-changing event. The last modification occurred at 28.645 minutes, leaving a 20.1-minute tail with no lasting change. Its 50 text-changing actions comprised 25 insertions and 25 removals arranged in 20 append--delete cycles. Early cycles added strings such as \texttt{ddd}; later cycles generally added individual letters. Every cycle returned to the completed text. The median cycle lasted 0.8395 seconds, and temporary characters were present for only 17.239 seconds, or 1.43\% of the tail. 

In essay \texttt{65825701}, 49 consecutive single-space input events occurred between 16.3479 and 24.7290 minutes. Each advanced the cursor by one character; the median interval between presses was 10.633 seconds. Replay confirmed that the word-token sequence was unchanged across this sequence. The later log also contained 62 left clicks after minute 25, whitespace edits, and a small deletion of the word \emph{either}. Its two paste events each inserted only a period and a space. 

The activity measure was standardized within each essay as
\[
 A_z=\frac{A-\operatorname{med}_e(A)}{1.4826\,\operatorname{MAD}_e(A)}.
\]
Prolonged low-activity tails made these denominators small, so earlier drafting intervals acquired extreme standardized values despite unexceptional raw activity ranks. Together, the two essays accounted for 45.76\% of within-essay squared variation in the state--activity interaction predictor in the whole corpus. 

We excluded these two essays from the  restoring-dynamics regressions and their associated resampling and activity-shift analyses, removing 58 windows and leaving 106,353 windows from 4,342 essays in the primary regression. The descriptive capture analysis retains all 4,781 exactly replayed essays, including these two: their exclusion concerned their influence on the activity regressions, rather than a reconstruction failure. Static finite-size and activity-distribution analyses also retain their original cohorts.

\subsection{Essay-level sources of the large-revision tail}
\label{app:large-revision-cases}

We inspected 12 selected essays contributing large complete revisions, using the saved episode partition and replaying the corresponding raw logs. The common cohort used for the size-stratified restoration analysis contained 292,009 eligible revisions from 4,241 essays. Its 301--700-character bin contained 288 revisions from 202 essays, including 144 with bulk operations (50.0\%); the 701+-character bin contained 51 revisions from 39 essays, including 39 with bulk operations (76.5\%). 

\begin{table*}[htbp]
\centering\small
\caption{Selected essay audits. $n_{>300}$ counts complete extracted revisions exceeding 300 gross changed characters in each log. $V_{\max}$, $N$, and $T$ describe the largest such revision: gross characters, native modifications, and duration in seconds. Descriptions summarize the inspected large episodes, not necessarily only the largest. The cases are selected examples, not an exhaustive ranking or a representative sample.}
\label{tab:large-revision-essays}
\setlength{\tabcolsep}{4pt}
\renewcommand{\arraystretch}{1.15}
\begin{tabularx}{\linewidth}{@{}lrrrrp{10cm}}
\toprule
Essay & $n_{>300}$ & $V_{\max}$ & $N$ & $T$ & Observed editing pattern\\
\midrule
\texttt{10768319} & 4 & 1,439 & 1 & 0 & Repetitive nonlexical strings inserted over several episodes, then removed in bulk.\\
\texttt{68216667} & 4 & 1,095 & 1,095 & 42.10 & Extended insertion and character-by-character removal of repetitive letter strings.\\
\texttt{23987928} & 4 & 518 & 332 & 133.19 & Insertion, replacement, and deletion of nonlexical keyboard sequences.\\
\texttt{30578676} & 3 & 836 & 1 & 0 & Nonlexical strings typed and removed, including a 471-character insertion followed by its bulk deletion.\\
\texttt{40319803} & 1 & 383 & 383 & 102.10 & Three-character deletion followed by 380 characters of on-topic prose.\\
\texttt{76169954} & 1 & 323 & 323 & 95.92 & Three-character deletion followed by 320 characters of on-topic prose.\\
\texttt{63108840} & 1 & 340 & 1 & 0 & Bulk removal of an outline containing planning headings.\\
\texttt{33301697} & 2 & 514 & 1 & 0 & Replacement of a prose passage by one character; another episode removes 324 characters of prose.\\
\texttt{22001544} & 1 & 334 & 1 & 0 & Bulk removal of a coherent autobiographical paragraph.\\
\texttt{13187058} & 1 & 1,031 & 1,031 & 328.32 & Character-by-character removal of a mixed-language personal passage.\\
\texttt{50786743} & 1 & 1,075 & 1 & 0 & Bulk removal of a coherent workplace example.\\
\texttt{67579301} & 1 & 3,031 & 1 & 0 & Replacement deleting 1,468 and inserting 1,563 characters, with substantial overlap between versions.\\
\bottomrule
\end{tabularx}
\end{table*}

The selected logs contain 24 complete revisions above 300 characters (Table~\ref{tab:large-revision-essays}). Two additional large candidates were incomplete under the extraction rule and omitted: a 1,125-character deletion in \texttt{68216667} and a 663-character replacement in \texttt{67579301}. 

The audit separates several sources of the tail. In \texttt{10768319}, three complete revisions of 647, 380, and 350 characters predominantly extended repetitive letter strings; a later single operation removed 1,439 characters and returned the document to its preceding prose ending. Similar sequences occurred in \texttt{68216667}, \texttt{23987928}, and \texttt{30578676}. Such activity is not limited to bulk edits: the 1,095-character deletion in \texttt{68216667} comprised 1,095 native modifications. Conversely, large deletions in \texttt{22001544} and \texttt{50786743} removed coherent prose, and \texttt{63108840} removed planning material. Bulk status therefore does not distinguish nonlexical activity from substantive editing.

Other cases clarify the operational meaning of episode size. In \texttt{40319803} and \texttt{76169954}, a small deletion initiated a revision episode containing substantial subsequent production. In \texttt{67579301}, the largest replacement changed document length by only 95 characters despite gross volume of 3,031, because both the removed and inserted passages contribute to $V$. 

These cases support retaining a distinction between the statistical revision-size tail and its behavioral interpretation. Large episodes include substantive passage revision, planning-text removal, extended production after a correction, and reversible nonlexical activity. The selected audit cannot establish the prevalence of these behaviors throughout the corpus, and no additional exclusions were made on its basis.

\section{Supplementary analysis of typing-activity scaling}
\label{app:typing-activity-scaling}

This appendix gives the distribution plots, fitting details, and threshold diagnostics supporting the main-text comparison of writing-activity measures. Temporal bursts, deletion runs, and revision/production episodes have different boundaries and units. We therefore assess their distributions separately rather than treating the fitted exponents as estimates of a universal quantity.

\subsection{Temporal bursts and deletion runs}
\label{app:temporal-burst-scaling}

For essay $e$, let $g_i$ be the interval between consecutive text-changing events and $m_e=\operatorname{med}(g_i)$. A temporal avalanche is a maximal sequence whose successive gaps satisfy $g_i\leq\delta_e$, with $\delta_e=1.25m_e$. Size $S$ is the number of constituent events, and duration is $T=(t_{\mathrm{last}}-t_{\mathrm{first}})/m_e$. Thus, $T$ is dimensionless and is distinct from revision-episode duration in seconds. Excluding the first and last avalanche of each essay to reduce session-boundary censoring yielded 5,270,050 primary avalanches across 4,992 essays; 3,019,125 had positive duration.

\begin{figure*}[htbp]
\centering
\includegraphics[width=.60\textwidth]{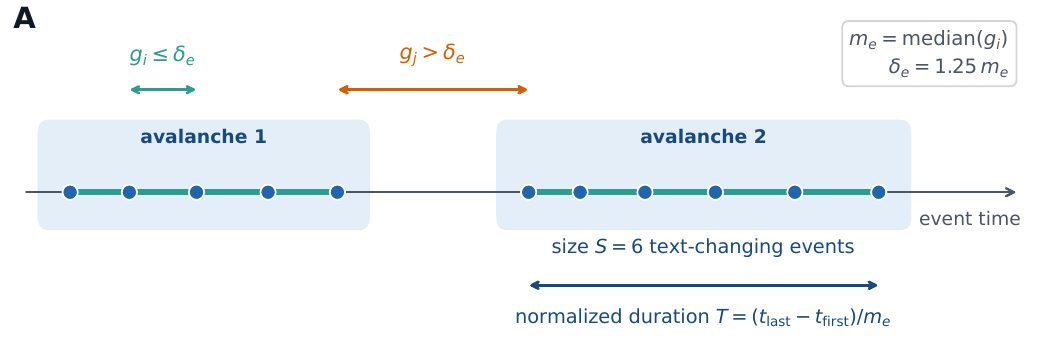}\par\vspace{3pt}
\includegraphics[width=.325\textwidth]{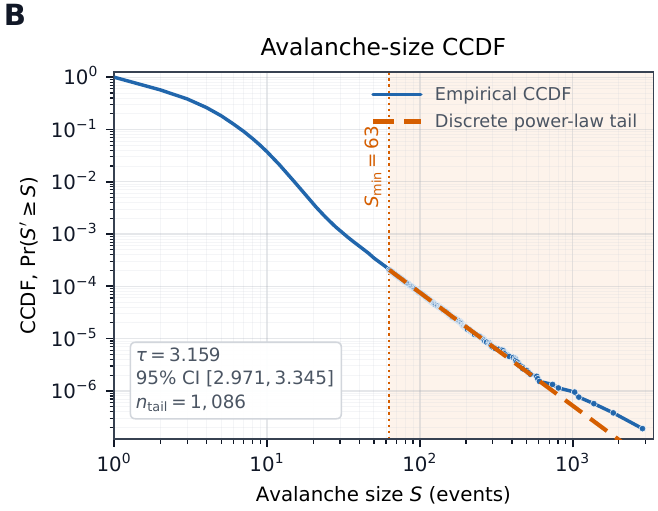}\hfill
\includegraphics[width=.325\textwidth]{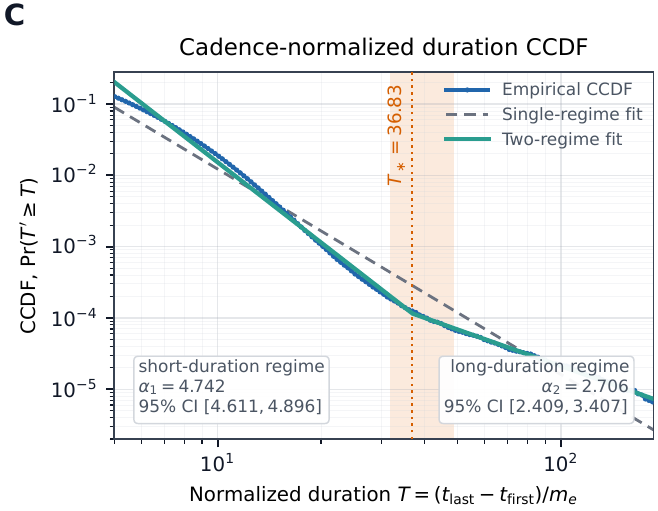}\hfill
\includegraphics[width=.325\textwidth]{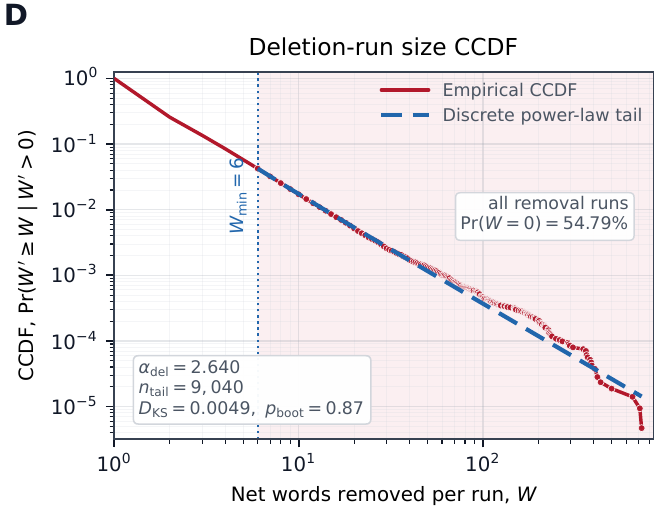}
\caption{\textbf{Temporal avalanches and deletion runs.} (A) Gap-based avalanche definition. (B) Empirical size CCDF and discrete power-law tail above $S_{\min}=63$, with probability-mass exponent 3.159. (C) Cadence-normalized duration CCDF; the descriptive two-regime fit has crossover $T_\ast=36.83$, with the shaded 95\% whole-essay bootstrap interval $[31.66,48.74]$. The dashed gray curve is the rejected single-regime fit. (D) Net word loss per deletion run, conditional on positive loss; the fitted tail begins at six words and has probability-mass exponent 2.640. The original saved panels are reproduced without refitting.}
\label{fig:typing-activity-scaling}
\end{figure*}

The avalanche-size tail above $S_{\min}=63$ contained 1,086 observations and gave $\hat\tau=3.159$ for $p(S)\propto S^{-\tau}$. Whole-essay resampling gave a median exponent of 3.170 and a 95\% interval of $[2.971,3.345]$. The reported bootstrap goodness-of-fit test did not reject the fitted tail ($p_{\mathrm{boot}}=1.00$). This is a non-rejection result, not the probability that the power-law hypothesis is true. The power law was strongly preferred to an exponential, while a lognormal remained a plausible competitor. These conclusions concern a small upper tail, not the entire size distribution.

A single continuous power law for durations above $T_{\min}=10.28$ was rejected. We therefore described the empirical CCDF, $\overline F_T(T)=\Pr(T'\geq T)$, by a continuous broken line in log--log coordinates:
\begin{equation}
\log\overline F_T(T)=
\begin{cases}
 a+b_1\log(T/T_\ast),&T\leq T_\ast,\\
 a+b_2\log(T/T_\ast),&T>T_\ast.
\end{cases}
\label{eq:app-duration-broken-line}
\end{equation}
The duration fit covered $5\leq T\leq185.85$. Its crossover was $T_\ast=36.83$, with 95\% whole-essay bootstrap interval $[31.66,48.74]$ and slopes $b_1=-3.742$ and $b_2=-1.706$. Their density-equivalent exponents, $1-b_1=4.742$ and $1-b_2=2.706$, describe the two fitted CCDF segments. This fit reduced log-CCDF residual sum of squares by 94.3\% relative to one line. The long-duration exponent was stable when the lower fitting boundary varied from 3 to 10, while the crossover ranged from 32.7 to 43.6. Avalanches near the crossover contained a median of 49 events. Because CCDF ordinates are dependent, the residual reduction is descriptive evidence of curvature, not an independent-observation significance test or proof of a distinct cognitive regime.

Deletion runs group consecutive removal or cut operations. Their size $W$ is the net decrease in document word count, rather than the number of deleted characters or actions. Of 471,452 runs, 258,321 (54.79\%) were word-count neutral, leaving 213,131 positive-loss runs. Above $W_{\min}=6$, the power-law estimate was $\hat\alpha_{\mathrm{del}}=2.640$, based on 9,040 runs (4.24\% of positive-loss runs). The fitted tail had $D_{\mathrm{KS}}=0.00493$, and the likelihood comparison favored the power law over an exponential. Deletion runs overlap behaviorally with other editing units; this is a complementary measurement, not an independent sample of writers.

Parameter uncertainty and goodness of fit used different procedures. The temporal-size exponent interval was obtained from 200 whole-essay bootstrap samples, refitting the exponent while holding the selected cutoff $S_{\min}=63$ fixed. This preserves dependence among bursts from the same essay, but does not propagate cutoff-selection uncertainty. The episode-size threshold scan separately used 500 whole-essay bootstrap samples at each specified threshold.

The original temporal-tail goodness-of-fit diagnostic instead attempted 100 independent-observation parametric simulations (base random seed 20260903). The initial fit selected a cutoff by minimizing the KS distance over at most 60 observed candidate values, subject to at least 1,000 tail observations. The goodness-of-fit routine first restricted the data to that selected tail, then reselected the observed cutoff and exponent using at most 40 candidates and a relaxed minimum count $\min(1000,\max(50,\lfloor n/10\rfloor))$, where $n$ is the input tail count. It generated $n$ independent observations from this internally refitted power law and repeated the cutoff and exponent selection on each simulated sample. Its 200,000-observation cap was inactive for both the size and duration fits.

The implementation calculated
\[
 \widehat p_{\mathrm{boot}}
 =\frac{1}{B_{\mathrm{ok}}}
   \sum_{b\in\mathcal B_{\mathrm{ok}}}
   \mathbf 1[D_b\geq D_{\mathrm{obs}}],
\]
without a plus-one adjustment, where $\mathcal B_{\mathrm{ok}}$
contains simulations whose refits completed successfully. The saved
size result was $1.00$, meaning that all retained simulated KS values
were at least as large as the internally recomputed observed value.
However, the saved summary did not retain $B_{\mathrm{ok}}$ or the
internally reselected observed cutoff, exponent, and KS statistic.
Consequently this value cannot yet be audited as a test of the
specifically reported $S_{\min}=63$ fit. The independent simulations
also omit within-essay dependence; the separate whole-essay exponent
bootstrap does not correct this limitation. We therefore treat the
saved result as an approximate diagnostic rather than confirmatory
goodness-of-fit evidence.

\subsection{Revision and production sizes}
\label{app:revision-production-scaling}

Following the revision-event approach of \citet{ConijnEtAl2024}, revisions begin with deletion or insertion into existing text. The adapted keystroke-only support-vector model selects endpoints within structurally bounded candidates, preserving native actions and the initial deletion core. Production episodes are maximal nonempty sequences of modifications outside extracted revisions. This partition uses neither sentence boundaries nor a pause threshold. The conservative complete subset excludes episodes reaching the recording boundary and revisions whose candidate interval reaches that boundary. The transferred endpoint model has not been independently validated against human revision annotations in KLiCKe.

Gross size is $V=V_{\mathrm{inserted}}+V_{\mathrm{deleted}}$, including whitespace and both components of replacements. It measures editing volume, not net document growth. We fitted discrete tails
\begin{equation}
 p(V=v\mid V\geq v_{\min})=
 \frac{v^{-\alpha}}{\zeta(\alpha,v_{\min})},\qquad v\geq v_{\min},
\label{eq:app-episode-discrete-tail}
\end{equation}
where $\zeta$ is the Hurwitz zeta function. For continuous power laws, the CCDF exponent is $\alpha-1$; for these discrete distributions this relationship is asymptotic. Thresholds were scanned from 100 to 500 characters. In each of 500 nonparametric whole-essay bootstrap samples, essays were sampled with replacement and their complete episode sets carried together into the refit. Exponents were refitted at each fixed threshold. The resulting pointwise percentile intervals account for essay clustering but not endpoint-model uncertainty or selection of a threshold after inspecting the scan. They are not goodness-of-fit tests.

\begin{figure*}[htbp]
\centering
\includegraphics[width=\textwidth]{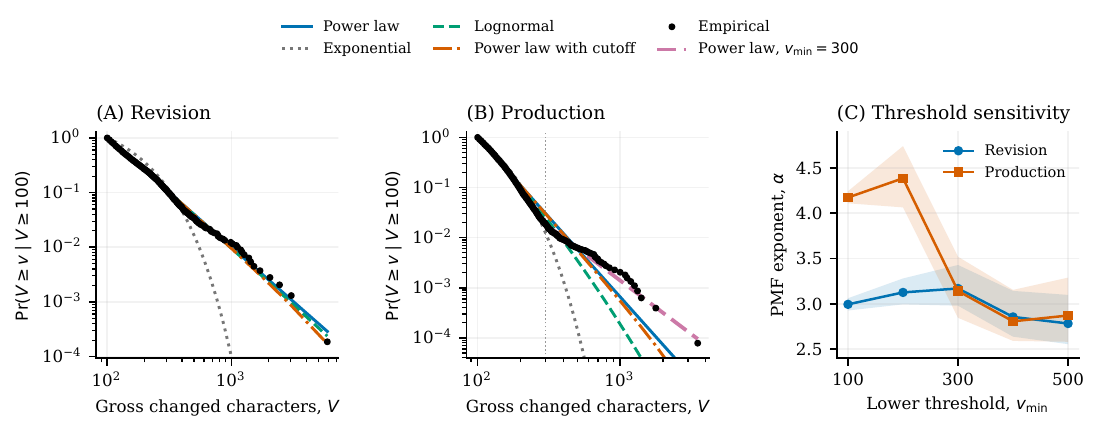}
\caption{\textbf{Revision and production size distributions.} (A,B) Empirical survival probabilities conditional on $V\geq100$ and four fitted distribution families. The additional production power law above 300 characters is plotted only on its support and multiplied by $236/12{,}742$ to share the conditional-on-100 probability scale. The guide at 300 is not an estimated breakpoint. Points use the saved subsampled empirical plotting grid. (C) Power-law exponents across fixed lower thresholds, with pointwise 95\% whole-essay bootstrap intervals from refitting. Bulk edits remain included throughout.}
\label{fig:revision-production-scaling}
\end{figure*}

Revision exponents remained near three across the scan (Table~\ref{tab:app-episode-scan}), ranging from 2.784 to 3.170. At 100 characters, $\hat\alpha=2.995$ with interval $[2.927,3.062]$, based on 5,352 revisions from 2,526 essays. Production exponents changed more markedly: 4.386 at 200 characters, 3.142 at 300, and 2.808 at 400. The apparent transition toward a heavier tail around 300--400 characters is therefore a candidate crossover rather than a fitted breakpoint. Support becomes sparse: only 236 production episodes from 187 essays remain above 300, and 116 from 89 essays above 400.

\begin{table*}[htbp]
\centering\small
\caption{Fixed-threshold gross-character fits, including bulk edits. Intervals are pointwise 95\% whole-essay bootstrap intervals; all ten rows had 500 valid refits. Counts at successive thresholds are nested and must not be added.}
\label{tab:app-episode-scan}
\begin{tabular}{llrrl}
\toprule
Episode type & $v_{\min}$ & Episodes & Essays & $\hat\alpha$ [95\% interval]\\
\midrule
Revision & 100 & 5,352 & 2,526 & 2.995 [2.927, 3.062]\\
Revision & 200 & 1,425 & 935 & 3.127 [2.995, 3.281]\\
Revision & 300 & 591 & 410 & 3.170 [2.980, 3.429]\\
Revision & 400 & 283 & 196 & 2.856 [2.638, 3.143]\\
Revision & 500 & 183 & 130 & 2.784 [2.556, 3.099]\\
Production & 100 & 12,742 & 3,709 & 4.174 [4.110, 4.243]\\
Production & 200 & 1,149 & 812 & 4.386 [4.065, 4.740]\\
Production & 300 & 236 & 187 & 3.142 [2.846, 3.520]\\
Production & 400 & 116 & 89 & 2.808 [2.591, 3.154]\\
Production & 500 & 80 & 61 & 2.872 [2.582, 3.290]\\
\bottomrule
\end{tabular}
\end{table*}

\subsection{Alternative models and sensitivity diagnostics}
\label{app:typing-model-comparisons}

Power laws were compared with exponential, lognormal, and exponentially truncated power-law alternatives on a common lower support for each comparison. In the episode analysis, models were trained on a fixed set of essays and evaluated on held-out essays. Test essays were then resampled to quantify uncertainty in the difference in predictive log likelihood, conditional on the trained models. This procedure differs from the exponent bootstrap above: it does not refit training models in each test-set resample. The likelihood contribution is per episode, so the more numerous observations close to the threshold strongly influence the comparison. A fit that appears closer at the most extreme observations need not have higher predictive likelihood over the entire selected tail.

Table~\ref{tab:app-tail-comparisons} reports selected comparisons. At 100 characters, revision sizes favored the power law over an exponential, while differences from the lognormal and cutoff alternatives were small and their intervals included zero. At the 400-character production threshold, the lognormal and cutoff alternatives outperformed the pure power law, although the test sample contained only 36 episodes. Neither a visually straight segment nor an approximately stable exponent establishes a unique distributional family. Likewise, non-rejection in the temporal-size and deletion-run goodness-of-fit tests cannot be transferred to revision or production episodes.

\begin{table*}[htbp]
\centering\small
\caption{Held-out comparisons for gross character volume. $\Delta\ell$ is mean test log likelihood per episode for the power law minus that for the named alternative; positive values favor the power law. Intervals resample test essays conditional on trained models. Only bulk-included primary fits are shown.}
\label{tab:app-tail-comparisons}
\begin{tabular}{llrlr}
\toprule
Episode type & $v_{\min}$ & Test episodes & Alternative & $\Delta\ell$ [95\% interval]\\
\midrule
Production & 400 & 36 & Exponential & -0.1117 [-0.2409, 0.0277]\\
Production & 400 & 36 & Lognormal & -0.0299 [-0.0495, -0.0082]\\
Production & 400 & 36 & Power law with cutoff & -0.0323 [-0.0529, -0.0086]\\
Revision & 100 & 1,598 & Exponential & 0.1942 [0.1152, 0.2835]\\
Revision & 100 & 1,598 & Lognormal & 0.0003 [-0.0006, 0.0011]\\
Revision & 100 & 1,598 & Power law with cutoff & 0.0006 [-0.0005, 0.0018]\\
\bottomrule
\end{tabular}
\end{table*}

\end{document}